\documentclass[runningheads]{llncs}

\usepackage{eccv}

\usepackage{eccvabbrv}

\usepackage{graphicx}
\usepackage{booktabs}

\usepackage[accsupp]{axessibility}  

\usepackage{hyperref}

\usepackage{orcidlink}
\usepackage{caption}
\usepackage{placeins}
\usepackage[table]{xcolor}
\usepackage{multirow}
\usepackage{float}
\usepackage{graphicx}
\usepackage[ruled,vlined]{algorithm2e}
\usepackage{amsmath,amssymb}
\usepackage{bm}

\begin{document}

\title{SFDATrack: Generalized Source-Free Domain Adaptive Tracking Under Adverse Weather Conditions} 

\titlerunning{Generalized Source-Free Domain Adaptive Tracking}




\author{
Siyuan Yao\inst{1}
\and
Ziqi Wang\inst{1,2}
\and
Ruiqi Yu\inst{3}
\and
Junqi Huang\inst{1}
\and
Wenqi Ren\inst{1}\textsuperscript{\dag}
\and
Xiaochun Cao\inst{1}\textsuperscript{\dag}
}

\authorrunning{S. Yao et al.}

\institute{
Shenzhen Campus of Sun Yat-sen University
\and
Beijing University of Posts and Telecommunications
\and
Nanyang Technological University\\[0.5em]
\email{
\{yaosiyuan04,zqwatcherbr0,yuruiqi422,huangjq3632,rwq.renwenqi\}@gmail.com,\\
caoxiaochun@mail.sysu.edu.cn
}
}

\maketitle
\begingroup
\renewcommand\thefootnote{}
\footnotetext{%
$^\dagger$ Corresponding authors.
}
\endgroup

\begin{abstract}
Domain adaptive visual object tracking under adverse weather conditions has garnered significant attention in recent years. Despite the impressive performance, existing methods heavily rely on the large-scale video frames from both source and target domains, which is impractical under rigid resource constraints where source data is unavailable. To overcome this limitation, we propose SFDATrack, a generalized source-free domain adaptive tracker that merely leverages adverse weather samples from the target domain for robust state estimation. Specifically, SFDATrack first employs a mean-teacher backbone with Dual Interactive Mamba (DIM) blocks to distill the candidate target tokens that are resilient to weather variations from classified, augmented samples. Afterwards, we introduce a hyperspherical prototype projection (HPP) module to project these tokens onto multi-domain prototypes within a latent hyperspherical space. By enforcing both domain-specific and domain-invariant properties of the multi-domain prototypes, SFDATrack can be seamlessly adapted to diverse weather conditions with powerful generalizability. Extensive experiments evaluated on various benchmarks demonstrate that SFDATrack achieves superior performance compared to state-of-the-art approaches. The code is available at \url{https://github.com/watcherBR0/sfdatrack}.
\keywords{Visual Object Tracking \and Source-Free Domain Adaptation \and Hyperspherical Embedding}
\end{abstract}    
\section{Introduction}
\label{sec:intro}

Visual object tracking (VOT) is an essential computer vision task, which aims to estimate the trajectory of an arbitrary target in video sequences under the guidance of first frame initialization. It serves as a critical component for a wide range of applications, spanning from autonomous driving, video surveillance and embodied intelligence. Driven by the powerful deep learning architectures, modern trackers have shown remarkable capability to predict the target state in controlled environments.

While the deep trackers have achieved impressive performance on standard benchmarks, a critical challenge persists for modern VOT, \emph{i.e.}, when a tracking model trained on a curated source domain (\emph{e.g.}, daytime) is deployed to an unseen target domain (\emph{e.g.}, nighttime or fog), it often suffers severe performance degradation. Such adverse weather conditions introduce complex, non-uniform perturbations like low contrast and atmospheric noise, which severely disrupt the model's identification ability to track the target object.

\begin{figure}[!t]
\centering

\includegraphics[ width=0.9\linewidth]{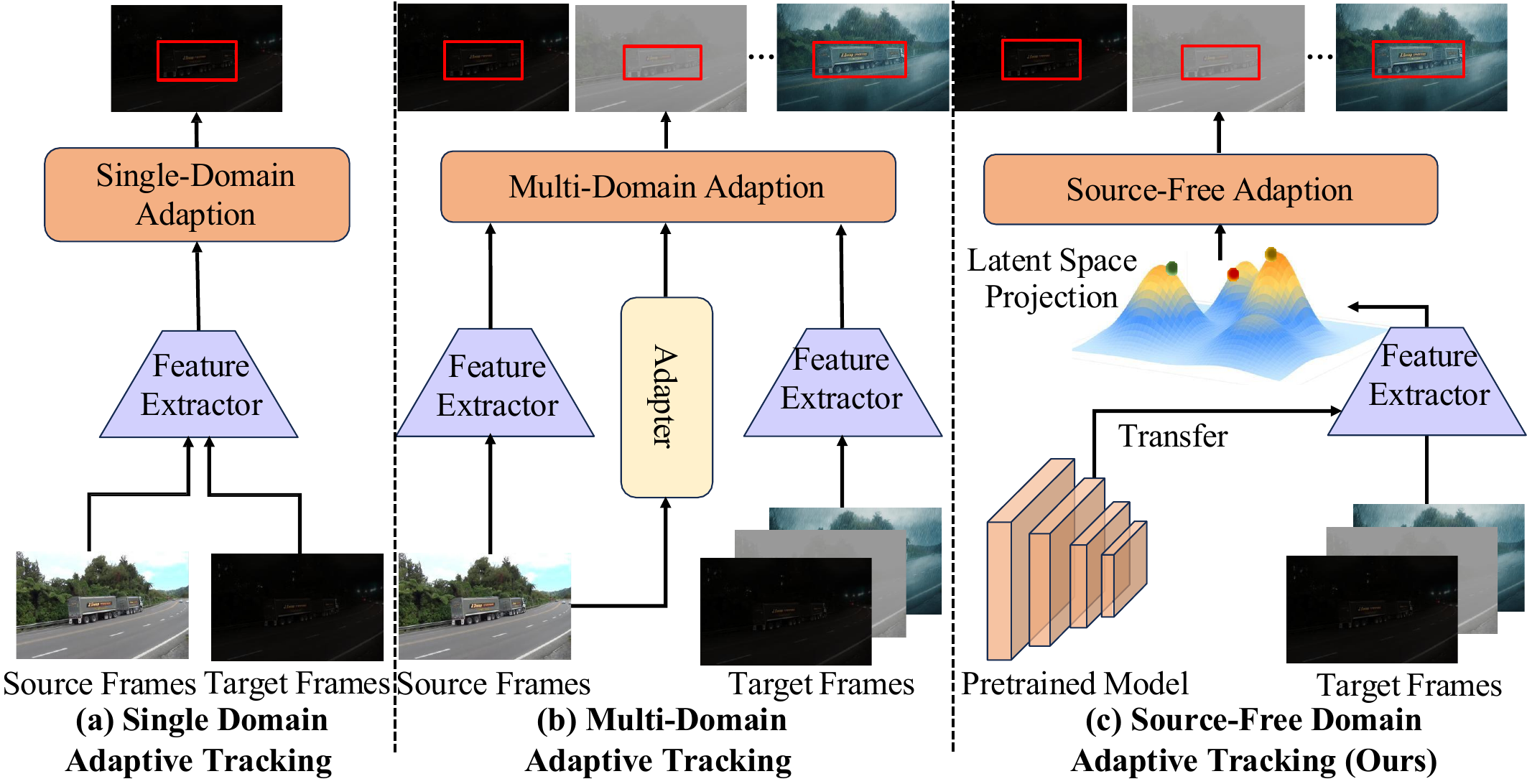}
\caption{Comparison of the domain adaptive tracking frameworks under adverse weather conditions. (a) Single domain adaptive tracking \cite{DBLP:conf/iros/Ye0ZCL21}. (b) Multi-domain adaptive tracking \cite{yao2025umdatrack}. (c) The proposed source-free domain adaptive tracker (SFDATrack). SFDATrack adapts a pre-trained model to multiple target domains via latent space projection, enabling seamless adaptation to diverse weather conditions without requiring any original source data.}
\label{fig:motivation}
\end{figure}

To overcome this challenge, recent studies have proposed Domain Adaptive Visual Object Tracking (DAVOT)~\cite{zhang2021domain,DBLP:conf/iros/Ye0ZCL21,DBLP:conf/iros/0001DY0LZ22,DBLP:conf/cvpr/Ye0ZPC22,DBLP:conf/aaai/WuYHWZZR25} as a promising solution. As shown in Fig.~\ref{fig:motivation}, the prevailing DAVOT methods typically follow an unsupervised single domain adaptation or multi-domain adaptation paradigms, which utilizes annotated source data and unlabeled target data to learn domain-invariant features across different weather conditions. However, these standard domain adaptive tracking framework depend on a critical yet often impractical assumption: the perpetual availability of the source domain data during adaptation. In real-world scenarios, this assumption is frequently invalid due to storage limitations, computational costs, or data privacy regulations. For instance, it is infeasible to collect large-scale source videos on a device with limited memory or to transmit them over bandwidth-constrained networks. This limitation raises a pivotal question: ‌\textit{\textbf{‌can a pre-trained source domain model be successfully adapted to arbitrary challenging target domains for object tracking without any access to the original source data?}}

In this paper, we propose SFDATrack, a novel source-free domain adaptive tracker that achieves robust visual tracking in adverse weather using only unlabeled target data. Our approach is built upon the idea that an object's essential visual semantics are resilient to weather variations in the embedding space, even as its low-level appearance shifts. Motivated by this, we design a mean-teacher network that performs multi-domain knowledge distillation for target state prediction. It comprises two novel components: 1) a Dual Interactive Mamba (DIM) module establishes bidirectional sequence interactions and facilitates context token exchange between differently augmented views of the target, effectively narrowing domain discrepancies. 2) a Hyperspherical Prototype Projection (HPP) module projects these tokens onto a set of multi-domain prototypes within a latent hyperspherical space. A dual alignment strategy is designed to minimize distribution differences among these prototypes, concurrently enforcing both domain-specific and domain-invariant properties for model generalization, enabling the tracker to adapt seamlessly across diverse weather conditions. To the best of our knowledge, this is the first work to address source-free domain adaptation in the VOT community.

In summary, the main contributions of this work are:

$\bullet$ We introduce SFDATrack, a novel framework that addresses the critical yet under-explored problem of source-free domain adaptive tracking under adverse weather conditions.

$\bullet$ We design a Dual Interactive Mamba (DIM) module to enhance the model's resilience to multiple weather conditions via bidirectional sequence interaction and context token exchange.

$\bullet$ We propose a Hyperspherical Prototype Projection (HPP) module to enforce both domain-specific and domain-invariant properties for model generalization. Extensive experiments demonstrate that SFDATrack achieves superior performance to existing state-of-the-art methods.

\section{Related Work}
\label{sec:related}

\subsection{Visual Object Tracking}
Visual object tracking (VOT) aims to localize target objects given the initial annotation throughout a video sequence. With the success of deep convolutional networks, Siamese-based trackers \cite{bertinetto2016fully,li2018high,he2018twofold,yao2021learning,Chen2022DTFSiam} have achieved significant progress. 
Following the success of Transformers in vision tasks, several works unified feature extraction and target localization into end-to-end attention frameworks. STARK \cite{DBLP:conf/iccv/0002PF0L21} proposed a spatio-temporal Transformer that directly regresses the target box via an encoder-decoder structure. MixFormer \cite{DBLP:conf/cvpr/CuiJ0W22} and OSTrack \cite{ye2022joint} further adopted single-stream architectures to jointly learn global context and target-specific representation, achieving state-of-the-art performance on LaSOT \cite{lasot} and GOT-10k \cite{got10k}. Recently, Mamba-based  models \cite{mamba,hatamizadeh2025mambavision} have introduced sequence-state modeling to overcome the quadratic complexity of attention and improve temporal modeling efficiency. MambaVLT \cite{liu2025mambavlt} introduces time-evolving multimodal state-space modeling for efficient vision-language tracking. MambaLCT \cite{li2025mambalct} employs a state-space model to dynamically update long-term contextual states. Despite this progress, most of these modern trackers are still trained and evaluated within a single domain, lacking sufficient generalization capability to unseen scenarios and suffering from significant performance degradation when facing domain shifts. To combat this, a series of domain adaptation (DA) tracking methods have been proposed. UDAT \cite{DBLP:conf/cvpr/Ye0ZPC22} develops a transformer-based unsupervised adaptation framework that bridges the feature discrepancy between daytime and nighttime domains. UMDATrack \cite{yao2025umdatrack} adopts a mean-teacher framework with domain-specific adapters to achieve unified multi-domain adaptation under adverse weather conditions. However, such domain-adaptive trackers still rely on access to source-domain samples during adaptation, which may not always be feasible.

\subsection{Source-Free Domain Adaptation}
In many real-world scenarios, sharing source data is infeasible due to privacy, copyright, and transmission constraints, while retaining complete source datasets on resource-constrained devices is also impractical. To overcome these limitations of traditional unsupervised domain adaptation (UDA), source-free domain adaptation (SFDA) \cite{DBLP:conf/icml/LiangHF20,DBLP:conf/kdd/ChidlovskiiCC16} has been proposed, which adapts a pretrained source model to a target domain without any access to source data, using only unlabeled target samples. Following \cite{DBLP:journals/pami/LiYDZS24}, existing methods can be grouped into data-centric and model-centric paradigms. Data-centric approaches compensate for the missing source by reconstructing or transforming target data, including image restoration and enhancement techniques under adverse conditions \cite{shi2022sanddustcnn}.  \textsc{SF(DA)}$^2$ \cite{HwangLSY24} constructs an augmentation graph to implicitly augment and cluster target features, thereby tightening the decision boundary, whereas \textsc{SF-YOLO} \cite{DBLP:conf/eccv/VarailhonAPG24} introduces a learnable target-specific augmentation network in the image space to transform target data as a surrogate for the missing source. In the model-centric paradigm, SFDA primarily relies on self-training methods such as a teacher-student framework to ensure robustness and stability. However, noisy pseudo labels from the teacher can mislead the student and eventually cause the whole system to collapse through EMA feedback. DRU \cite{DBLP:conf/eccv/KhanhNPTJ24} dynamically retrains the student and selectively updates the EMA teacher based on the student's evolution. LPLD \cite{DBLP:conf/eccv/YoonKPKJS24} distills knowledge from low-confidence predictions to enhance model generalization. While prior SFDA studies focus on static recognition tasks, tracking introduces unique challenges: predictions must remain temporally coherent, and noisy pseudo labels that accumulate across frames will cause drift.

\section{Method}
\label{sec:method}

In this section, we present the overall architecture of SFDATrack. The whole model follows a mean-teacher pipeline, which conducts multiple-domains knowledge distillation for target object localization. SFDATrack consists of two crucial components: a Dual Interactive Mamba (DIM) module and a Hyperspherical Prototype Projection (HPP) module.

\begin{figure*}[t]
  \centering
  \includegraphics[width=\textwidth]{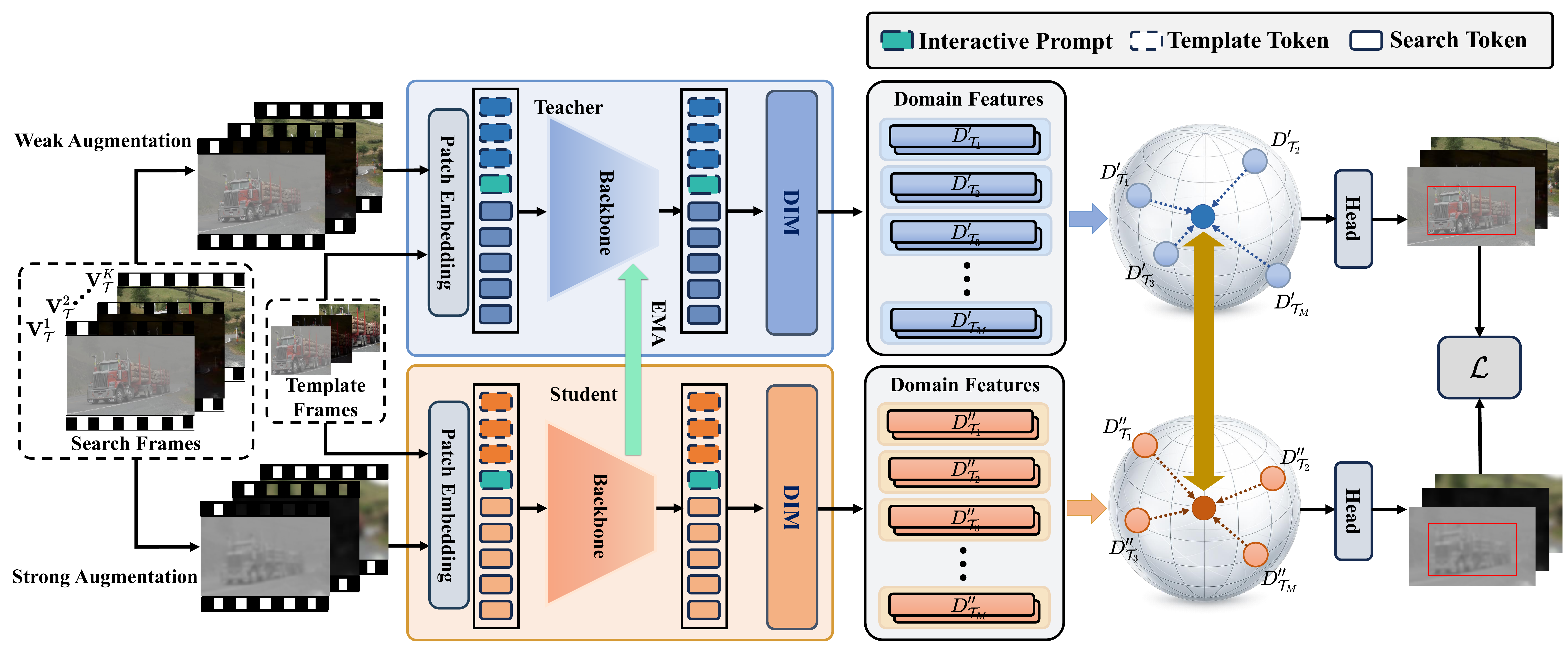}
  \caption{ Overview of the proposed SFDATrack. The input target video is first augmented into weak and strong views, which are processed by the teacher and student encoders, respectively. The extracted features are further fed into the Dual Interactive Mamba (DIM) module and Hyperspherical Prototype Projection (HPP) module to enhance domain consistency and representation robustness. Finally, a total loss $\mathcal{L}$ is applied to jointly optimize the tracking framework, ensuring stable and consistent feature learning across domains.
  }
  \label{fig:framework}
\end{figure*}

\subsection{Problem Definition}
\label{sec:problem}

Formally, let $D_S$ denotes the source domain data and $\mathcal{M}_{\mathcal{S}}$ is the source domain pretrained model, the set of data from the target domains are denoted as $\mathcal{D}_{\mathcal{T}}=\left \{ D_{\mathcal{T}_{1}},D_{\mathcal{T}_{2}},\cdots, D_{\mathcal{T}_{M}} \right \}$, where $\mathcal{D}_{\mathcal{T}_{i}}=\left \{ x_{\mathcal{T}_{i}}^{j} \right \}_{j=1}^{N_{\mathcal{T}_{i}}}$ is the $i$-th target domain and $x_{\mathcal{T}_{i}}^{j}$ is the corresponding $j$-th image, $M$ and $N_{\mathcal{T}_{i}}$ denote the number of target domains and target domain samples, respectively. In the source-free setting, the source data $D_S$ are inaccessible during adaptation, and only the pre-trained source model $\mathcal{M}_{S}$ is available. Our goal is to adapt the pretrained tracking model $\mathcal{M}_{S}$ to the set of target domains  $\mathcal{D}_{\mathcal{T}}$.  

As shown in Fig.~\ref{fig:framework}, SFDATrack follows a mean teacher pipeline to achieve source-free domain adaptation. Given $K$ videos $\mathbb{V}_{\mathcal{T}}=\left \{\mathbf{V}_{\mathcal{T}}^{1}, \mathbf{V}_{\mathcal{T}}^{2},\cdots,\mathbf{V}_{\mathcal{T}}^{K}  \right \}$ from the target domains, we simultaneously perform weak and strong augmentations on the video frames. The teacher branch generates pseudo-labels from the weakly augmented video frames, while the student branch receives another strongly augmented version. Both networks share the same architecture but have different parameter sets $\theta^{\mathcal{T}}$ and $\theta^{\mathcal{S}}$. The teacher's parameters are updated through an Exponential Moving Average (EMA), \emph{i.e.}, $\theta^{\mathcal{T}} \leftarrow \alpha\,\theta^{\mathcal{T}} + (1-\alpha)\,\theta^{\mathcal{S}}$, where $\alpha$ is the momentum coefficient controlling the update rate of the teacher model. 

\begin{figure}[t]
  \centering
  \includegraphics[width=4.2in]{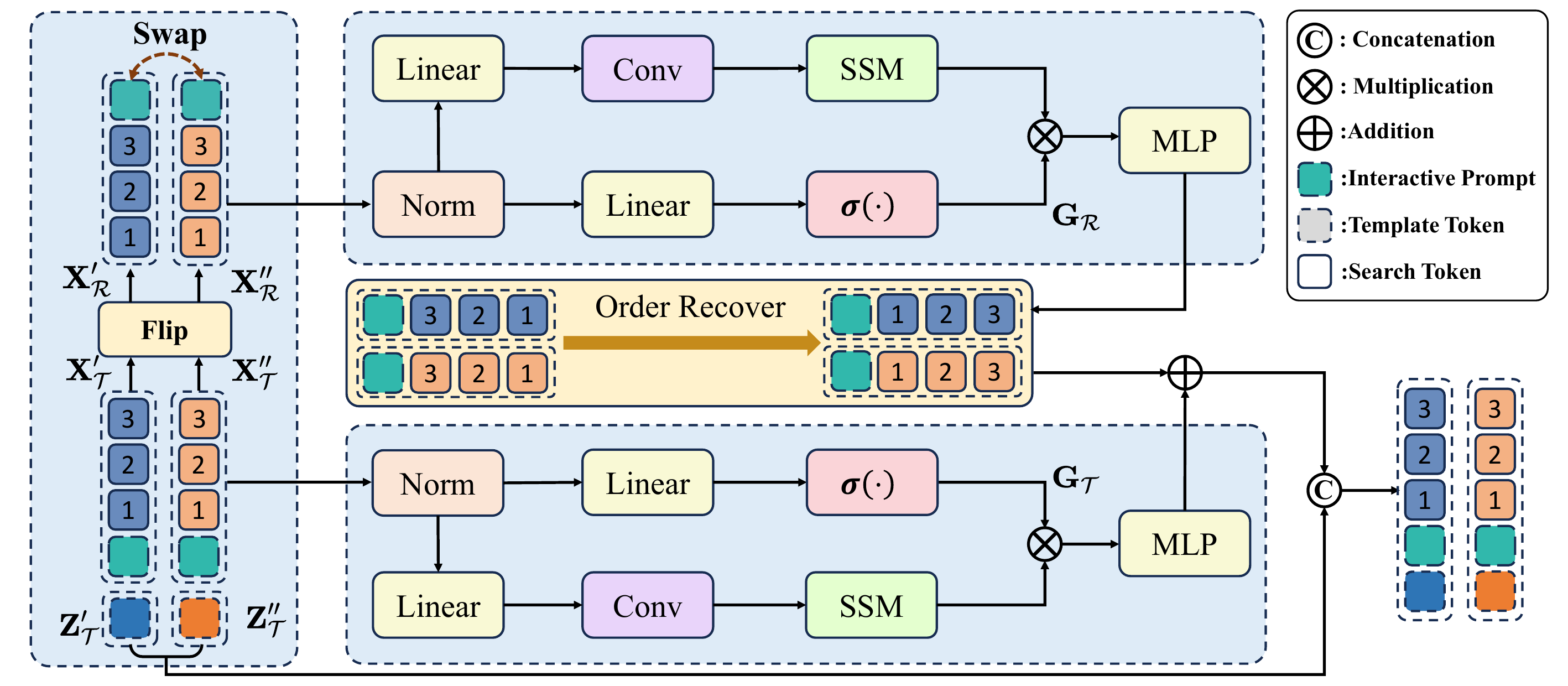}
  \caption{ Illustration of the Dual Interactive Mamba (DIM) module. The DIM module processes feature tokens from two different augmentations through bidirectional Mamba blocks with prompt tokens exchange. For simplicity, all prime and double-prime notations (e.g., $\mathbf{G}_{\mathcal{T}}^{\prime}$, $\mathbf{G}_{\mathcal{R}}^{\prime\prime}$) are omitted in the figure.
}
  \label{fig:DIM}
\end{figure}

\subsection{Dual Interactive Mamba}
\label{sec:mamba}
Although the generic mean-teacher framework provides an efficient way for domain knowledge transfer, their effectiveness and generalizability in multiple weather conditions are greatly limited due to the significant discrepancies of the different target domains. To address the issue, we introduce a Dual Interactive Mamba (DIM) block to distill the candidate target tokens that are resilient to weather variations from classified, augmented samples.   

Specifically, suppose we have obtained the encoded features $\mathbf{F}_{\mathcal{T}}^{\prime}=\left [ \mathbf{Z}_{\mathcal{T}}^{\prime},\mathbf{X}_{\mathcal{T}}^{\prime} \right ]$ and $\mathbf{F}_{\mathcal{T}}^{\prime\prime}=\left [ \mathbf{Z}_{\mathcal{T}}^{\prime\prime},\mathbf{X}_{\mathcal{T}}^{\prime\prime} \right ]$ in the teacher and student networks, where $\mathbf{Z}_{\mathcal{T}}^{\prime}$, $\mathbf{Z}_{\mathcal{T}}^{\prime\prime} \in\mathbb{R}^{L_{Z}\times C}$ denote the template tokens, $\mathbf{X}_{\mathcal{T}}^{\prime}$, $\mathbf{X}_{\mathcal{T}}^{\prime\prime} \in\mathbb{R}^{L_{X}\times C}$ denote the candidate search tokens, $L_{Z}$, $L_{X}$ and $C$ are length and feature dimension of the template-search tokens, respectively. Note that the weak and strong augmented branches share the similar target object's visual semantics, we introduce a set of learnable interactive prompt tokens $\mathbf{P}_{\mathcal{T}}^{\prime} $ and  $\mathbf{P}_{\mathcal{T}}^{\prime\prime}$ to establish bidirectional sequence interactions and facilitates context token exchange. As shown in Fig. \ref{fig:DIM}, we flip the dual search tokens and concatenate them with the interactive prompts for knowledge exchange as follows:


\begin{equation}
\begin{aligned}
\mathbf{X}_{\mathcal{R}}^{\prime} &= \mathrm{Concat}(\mathbf{P}^{\prime\prime};\mathcal{R}(\mathbf{X}_{\mathcal{T}}^{\prime})), \\
\mathbf{X}_{\mathcal{R}}^{\prime\prime} &= \mathrm{Concat}(\mathbf{P}^{\prime};\mathcal{R}(\mathbf{X}_{\mathcal{T}}^{\prime\prime})).
\end{aligned}
\end{equation}
where $\mathcal{R}(\cdot)$ denotes the reverse function of the search tokens. Then we normalize and linearly project $\mathbf{X}_{\mathcal{T}}^{\prime}$ and $\mathbf{X}_{\mathcal{R}}^{\prime}$ to obtain intermediate token representations. These representations are then passed through a sigmoid function to generate the gating vectors $\mathbf{G}_{\mathcal{T}}^{\prime}$ and $\mathbf{G}_{\mathcal{R}}^{\prime}$. The same process is applied to the corresponding tokens in the second branch to obtain $\mathbf{G}_{\mathcal{T}}^{\prime\prime}$ and $\mathbf{G}_{\mathcal{R}}^{\prime\prime}$, then we fed these tokens into a shared Mamba block, which can be given by:

\begin{equation}
\begin{aligned}
\mathbf{Y}_{\mathcal{T}}^{\prime} &= \mathcal{SSM}(\psi(\mathbf{X}_{\mathcal{T}}^{\prime})), \mathbf{Y}_{\mathcal{R}}^{\prime} &= \mathcal{SSM}(\psi(\mathbf{X}_{\mathcal{R}}^{\prime})).
\end{aligned}
\end{equation}
where $\psi(\cdot)$ performs channel mixing through a point-wise convolution and a SiLU activation to refine feature channels and $\mathcal{SSM}(\cdot)$ denotes the Mamba state-space operator. Similarly, the second branch follows the same procedure to produce $\mathbf{Y}_{\mathcal{T}}^{\prime\prime}$ and $\mathbf{Y}_{\mathcal{R}}^{\prime\prime}$. Finally, we fuse the gated features from both branches to obtain the enhanced search tokens:


\begin{equation}
\begin{aligned}
\mathbf{X}_{\mathcal{T}}^{\prime}=\mathcal{MLP}(\mathbf{G}_{\mathcal{T}}^{\prime}\odot \mathbf{Y}_{\mathcal{T}}^{\prime}) + \mathcal{R}(\mathcal{MLP}(\mathbf{G}_{\mathcal{R}}^{\prime}\odot \mathbf{Y}_{\mathcal{R}}^{\prime}))), \\
\mathbf{X}_{\mathcal{T}}^{\prime\prime}=\mathcal{MLP}(\mathbf{G}_{\mathcal{T}}^{\prime\prime}\odot \mathbf{Y}_{\mathcal{T}}^{\prime\prime}) + \mathcal{R}(\mathcal{MLP}(\mathbf{G}_{\mathcal{R}}^{\prime\prime}\odot \mathbf{Y}_{\mathcal{R}}^{\prime\prime}))), \\
\end{aligned}
\end{equation}
By stacking multiple Dual Interactive Mamba blocks, the model progressively enhances the cross-view semantic alignment and strengthens the bidirectional information flow between the weak and strong augmentations.

\subsection{Hyperspherical Prototype Projection}
\label{sec:prototype}
As the source domain data is not accessible in the adaptation process, distilling the original pretrained model into different target domains inevitably suffers significant performance degradation. To address this issue, we propose a \textit{Hyperspherical Prototype Projection} (HPP) module to enhance generalizability. Specifically, we perform average pooling on the output search features $\mathbf{X}_{\mathcal{T}}^{\prime}$ of DIM  to be $\mathbf{X}_{\mathcal{T}}^{\prime} = [\mathbf{x}_1, \mathbf{x}_2, \ldots, \mathbf{x}_N]^\top \in \mathbb{R}^{N \times D}$ and project these features onto a learnable prototype space $\mathcal{P} = \{ \mathbf{p}_k \}_{k=1}^{K}$~\cite{DBLP:journals/tip/YaoGYRC25}, which can be given by:

\begin{equation}
s_{i,k} =
\frac{\exp\!\left(\frac{1}{\tau}\,\mathbf{x}_i^\top \mathbf{p}_k\right)}
{\sum_{k'} \exp\!\left(\frac{1}{\tau}\,\mathbf{x}_i^\top \mathbf{p}_{k'}\right)},
\label{eq:proto_proj}
\end{equation}
where $\tau$ is a temperature parameter. Thus we can construct a prototype similarity matrix $\mathbf{S} = \left \{ s_{i,k} \right \}_{i=1,k=1}^{N,K} \in \mathbb{R}^{N\times K}$, where each row corresponds to the similarity distribution of a feature to all prototypes. We define a cost map $\mathbf{C}$ to measure the dissimilarity between features and prototypes as:
\begin{equation}
  \mathbf{C}_{i,k} = 1 - \frac{\mathbf{x}_i^\top \mathbf{p}_k}{\|\mathbf{x}_i\|_2 \, \|\mathbf{p}_k\|_2}.
  \label{eq:cost}
\end{equation}

As it's not trivial to learn the prototype space representation, similar to \cite{caron2020unsupervised}, we introduce a soft assignment matrix $\mathbf{Q} \in \mathbb{R}^{N \times K}$ that enforces the features can be projected to the unit hypersphere as below:
\begin{equation}
  \mathcal{Q} = \left\{ \mathbf{Q} \in \mathbb{R}_+^{N \times K} \mid 
  \mathbf{Q}\mathbf{1}_K = \tfrac{1}{N}\mathbf{1}_{N}, \;
  \mathbf{Q}^\top \mathbf{1}_{N} = \tfrac{1}{K}\mathbf{1}_K \right\},
  \label{eq:uq}
\end{equation}
where $\mathbf{1}_K$ and $\mathbf{1}_N$ denote all-one vectors of dimension $K$ and $N$, respectively. This constraint ensures that each prototype can be equally assigned within a mini batch. The soft assignment matrix $\mathbf{Q}$ is obtained by minimizing the assignment cost and encourages smooth, balanced distributions:
\begin{equation}
\min_{\mathbf{Q} \in \mathcal{Q}} \langle \mathbf{Q}, \mathbf{C} \rangle - \varepsilon \, \mathcal{H}(\mathbf{Q}),
\label{eq:ot}
\end{equation}
where $\mathcal{H}(\mathbf{Q}) = - \sum_{i,k} \mathbf{Q}_{i,k} \log \mathbf{Q}_{i,k}$ is the entropy term, and $\varepsilon$ controls the degree of assignment smoothness.
This optimization problem is efficiently solved via the Sinkhorn-Knopp algorithm~\cite{DBLP:conf/NeurIPS/Cuturi13}.

\noindent\textbf{Domain-Specific Alignment.} For multi-domain adaptation tracking task, the augmented samples related to the same weather in both teacher-student networks should maintain the uniform localization capabilities. To achieve this, we propose a Domain-Specific Alignment (DSA) strategy to ensure the localization consistency. Suppose the augmented samples are collected from $M$ domains, as described in Eq.~\ref{eq:proto_proj}, we send them into the student network and construct $M$ domain-specific similarity matrix $\{\mathbf{S}_{m}^{*} \in \mathbb{R}^{N \times K}\}_{m=1}^{M}$ in the prototype space. Correspondingly, the teacher provides the soft assignment targets $\{\mathbf{Q}_{m}^{*}\in\mathbb{R}^{N \times K}\}_{m=1}^{M}$ as pseudo labels for $M$ domains. We adopt a domain-specific loss function to align each student prediction with its corresponding teacher model at all of the weather domains:
\begin{equation}
  \mathcal{L}_{\text{DSA}} =
  -\frac{1}{MNK}
  \sum_{m=1}^{M}
  \sum_{i=1}^{N}
  \sum_{k=1}^{K}
  \mathbf{Q}_{m,i,k}^{*} \log \mathbf{S}_{m,i,k}^{*}.
  \label{eq:intra_domain_loss}
\end{equation}

\noindent\textbf{Domain-Invariant Alignment.} We propose a Domain-Invariant Alignment (DIA) strategy to enforce the consistency of multiple domains' representation in the latent hyperspherical space. Specifically, we first perform \textit{k}-means clustering on the encoded DIM output $\mathbf{X}_{\mathcal{T}}^{\prime} = [\mathbf{x}_1, \mathbf{x}_2, \ldots, \mathbf{x}_N]^\top \in \mathbb{R}^{N \times D}$ to be $M$ clusters
$\mathbf{V} = [\mathbf{v}_1, \dots, \mathbf{v}_M]^\top \in \mathbb{R}^{M \times D}$, we calculate the normalized clustering weight $\mathbf{W} = \left \{ w_{i,m} \right \}_{i=1,m=1}^{N,M} $ by: 
\begin{equation}
\centering
\begin{aligned}
w_{i,m} = 
\frac{\big(1 + \|\mathbf{x}_i - \mathbf{v}_m\|^2\big)^{-1}}
{\sum_{j} \big(1 + \|\mathbf{x}_i - \mathbf{v}_{j}\|^2\big)^{-1}},
\end{aligned}
\label{eq:soft-weights}
\end{equation}
where $\mathbf{v}_m$ is the $m$-th clustered center, $w_{i,m}$ indicates the weight of the $i$-th sample belonging to the $m$-th domain. 
Then we construct a clustered domain-invariant representation $\mathbf{X}^{*} = \mathrm{MLP}(\mathbf{W}\mathbf{V}) \in \mathbb{R}^{N \times D}$ by fusing the cluster-weighted features from different domains using a lightweight multilayer perceptron. 

Similarly to DSA, we project the clustered domain-invariant feature $\mathbf{X}^{*}$ into latent hyperspherical space~\cite{DBLP:conf/nips/MettesPS19,DBLP:journals/tip/YaoSXWC24}, and construct a domain-invariant similarity matrix $\hat{\mathbf{S}}^{*} \in \mathbb{R}^{N \times K}$. The teacher network generates the soft assignment targets $\mathbf{Q}^{*}\in\mathbb{R}^{N \times K}$. The domain-invariant alignment loss is defined as:
\begin{equation}
\mathcal{L}_{\text{DIA}} = 
\mathrm{KL}(\mathbf{Q}^{*} \| \mathbf{S}^{*})
= \sum_{i=1}^{N} \sum_{k=1}^{K} 
\mathbf{Q}_{i,k}^{*}
\log \frac{\mathbf{Q}_{i,k}^{*}}{\mathbf{S}_{i,k}^{*}}.
\label{eq:kl-global}
\end{equation}
The loss function in Eq. \ref{eq:kl-global} enforces the distribution alignment of the clustered domain-invariant features in teacher-student networks, allowing the tracker to alleviate the model biases to a specific domain.

\begin{table*}[t]
\centering
\caption{ Performance comparisons with state-of-the-art trackers on synthetic datasets: GOT-10k-Foggy, DTB70-Foggy, GOT-10k-Dark, DTB70-Dark, GOT-10k-Rainy and DTB70-Rainy. The top three results are highlighted with \textcolor{red}{red}, \textcolor{blue}{blue} and \textcolor{green}{green} fonts, respectively. Trackers above the double line denote \textbf{generic} methods, and those below denote \textbf{cross-domain} methods.}
\label{tab:comparison}
\small
\setlength{\tabcolsep}{3pt} 
\resizebox{\columnwidth}{!}{
\begin{tabular}{@{}c|ccc|cc|ccc|cc|ccc|cc@{}}
\toprule
\multirow{2.5}{*}{Tracker} & \multicolumn{3}{c|}{GOT-10k-Foggy} & \multicolumn{2}{c|}{DTB70-Foggy} & \multicolumn{3}{c|}{GOT-10k-Dark} & \multicolumn{2}{c|}{DTB70-Dark} & \multicolumn{3}{c|}{GOT-10k-Rainy} & \multicolumn{2}{c}{DTB70-Rainy} \\
\cmidrule(lr){2-16}
& AO & SR$_{0.50}$ & SR$_{0.75}$ & AUC & P & AO & SR$_{0.50}$ & SR$_{0.75}$ & AUC & P & AO & SR$_{0.50}$ & SR$_{0.75}$ & AUC & P \\
\midrule

SiamRPN++\cite{li2019siamrpn++} & 58.4 & 64.9 & 51.2 & 55.80 & 74.70 & 56.6 & 60.8 & 45.1 & 48.80 & 70.30 & 56.2 & 61.4 & 46.8 & 51.52 & 71.96 \\

DiMP\cite{bhat2019learning} & 57.6 & 64.2 & 50.4 & 53.80 & 69.50 & 56.9 & 60.4 & 44.3 & 55.20 & 72.30 & 57.9 & 63.8 & 49.2 & 57.32 & 75.21 \\

AVTrack\cite{lilearning} & 56.9 & 63.5 & 49.5 & 52.35 & 68.09 & 55.3 & 62.3 & 46.2 & 56.66 & 72.21 & 57.5 & 63.4 & 48.1 & 60.21 & 79.53 \\

OSTrack\cite{ye2022joint} & 61.9 & 71.7 & 59.7 & 57.54 & 73.95 & 61.3 & 70.9 & 51.5 & 59.23 & 77.43 & 61.6 & 71.0 & 58.6 & 63.16 & 83.60 \\

ROMTrack\cite{Cai_2023_ICCV} & 63.6 & 70.9 & 56.7 & 59.05 & 76.59 & 60.8 & 71.1 & 51.7 & 60.80 & 77.95 & 62.7 & 73.4 & 60.1 & 63.21 & 83.25 \\

AQATrack\cite{xie2024autoregressive} & 64.9 & 72.8 & 59.7 & 57.28 & 75.61 & 61.7 & 70.6 & 52.5 & 61.17 & 79.87 & 63.4 & 72.3 & \textcolor{green}{61.8} & 63.12 & 83.55 \\

SeqTrack\cite{chen2023seqtrack} & \textcolor{green}{65.2} & \textcolor{green}{74.6} & 56.3 & 60.21 & 78.70 & 61.4 & 70.5 & 52.3 & 62.84 & 81.57 & 65.1 & 75.0 & 60.3 & 63.75 & 83.28 \\

DropTrack\cite{dropmae2023} & 64.9 & 73.8 & 58.5 & 59.95 & 77.66 & 62.2 & 72.5 & 54.3 & 61.98 & 80.21 & 65.3 & 75.3 & 60.4 & 62.87 & 83.13 \\

EVPTrack\cite{shi2024evptrack} & 63.5 & 70.7 & 56.5 & 57.96 & 75.45 & 62.7 & 71.8 & 53.9 & 63.01 & 81.12 & 65.5 & 75.2 & 60.5 & 64.03 & 84.11 \\

ARTrackV2\cite{Bai_2024_CVPR} & 64.8 & 73.0 & \textcolor{green}{59.9} & 62.25 & 80.15 & 63.1 & 72.8 & 53.9 & 62.87 & 80.56 & 66.2 & 75.8 & 61.2 & 63.84 & 83.32 \\

UncTrack\cite{DBLP:journals/tip/YaoGYRC25} & 63.9 & 72.4 & 59.0 & 59.02 & 76.79 & 63.5 & 72.8 & 56.2 & 59.72 & 77.14 & 63.8 & 71.8 & 57.6 & 65.09 & 84.57 \\

ORTrack\cite{Wu_2025_CVPR} & 60.7 & 70.0 & 51.2 & 54.94 & 71.92 & 56.9 & 67.2 & 43.4 & 56.29 & 71.75 & 63.3 & 72.6 & 55.7 & 65.40 & \textcolor{green}{84.70} \\

SGLATrack\cite{DBLP:conf/cvpr/XueZLZLXS25} & 62.4 & 72.7 & 54.4 & 55.28 & 71.64 & 56.0	& 65.9 & 43.8 & 58.79 & 74.97 & 63.8 & 73.8 & 57.5 & 63.93 & 83.30 \\

UETrack\cite{Kang_2026_CVPR} & 63.1 & 71.9 & 56.5 & 59.19 & 74.64 & \textcolor{green}{65.3} & \textcolor{green}{75.1} & \textcolor{blue}{58.0} & 60.95 & 77.02 & 66.3 & 75.7 & 59.8 & \textcolor{green}{66.73} & 84.42 \\

\midrule
\midrule

MLKD-Track\cite{DBLP:journals/corr/abs-2312-07884} & 52.3 & 62.3 & 49.1 & 52.46 & 70.32 & 53.8 & 61.6 & 46.9 & 55.21 & 73.68 & 57.3 & 64.8 & 57.1 & 56.89 & 74.12 \\

SAM-DA\cite{Yao2023SAMDA} & 50.2 & 60.5 & 48.3 & 51.33 & 69.89 & 55.4 & 63.1 & 48.3 & 57.15 & 75.12 & 60.2 & 66.1 & 57.6 & 57.63 & 76.12 \\

UDAT-CAR\cite{DBLP:conf/cvpr/Ye0ZPC22} & 51.5 & 60.3 & 45.2 & 50.21 & 69.41 & 56.8 & 64.2 & 49.1 & 57.20 & 75.80 & 59.5 & 65.2 & 55.3 & 56.42 & 75.36 \\

DCPT\cite{DBLP:conf/icra/ZhuTCH0QLL24} & 61.6 & 70.2 & 56.9 & 58.31 & 75.33 & 62.4 & 70.5 & 54.2 & 61.87 & 80.11 & 62.3 & 70.1 & 59.8 & 61.68 & 82.56 \\

LVPTrack\cite{DBLP:conf/aaai/WuYHWZZR25} & 64.5 & 74.0 & 58.4 & \textcolor{green}{63.51} & \textcolor{green}{83.33} & 62.2 & 72.6 & 53.4 & \textcolor{green}{64.29} & \textcolor{green}{82.36} & \textcolor{green}{67.9} & \textcolor{green}{77.0} & 60.6 & 65.32 & 84.65 \\

UMDATrack\cite{yao2025umdatrack} & \textcolor{blue}{66.6} & \textcolor{blue}{75.8} & \textcolor{blue}{62.2} & \textcolor{red}{66.21} & \textcolor{blue}{86.05} & \textcolor{blue}{65.4} & \textcolor{blue}{75.3} & \textcolor{green}{57.3} & \textcolor{blue}{66.07} & \textcolor{blue}{85.72} & \textcolor{blue}{68.5} & \textcolor{blue}{78.4} & \textcolor{blue}{63.2} & \textcolor{blue}{66.75} & \textcolor{blue}{87.60} \\

\rowcolor{gray!20} \textbf{SFDATrack} & \textcolor{red}{70.1} & \textcolor{red}{80.6} & \textcolor{red}{67.1} & \textcolor{blue}{65.58} & \textcolor{red}{86.47} & \textcolor{red}{67.3} & \textcolor{red}{78.0} & \textcolor{red}{62.2} & \textcolor{red}{66.18} & \textcolor{red}{85.96} & \textcolor{red}{72.1} & \textcolor{red}{82.6} & \textcolor{red}{70.1} & \textcolor{red}{67.22} & \textcolor{red}{88.32}  \\

\bottomrule
\end{tabular}}
\end{table*}

To train SFDATrack, we integrate the target supervision loss, the domian-specific and domain-invariant alignment loss to achieve robust source-free domain adaptation. The target supervision loss $\mathcal{L}_S$ consists of the classification, localization L1, and generalized IoU terms following \cite{ye2022joint}:
\begin{equation}
\mathcal{L}_S = \mathcal{L}_{CLS} + \beta\,\mathcal{L}_{1} + \gamma\,\mathcal{L}_{GIoU},
\end{equation}
The whole training loss is formulated as:
\begin{equation}
\mathcal{L} = \mathcal{L}_S + \omega\,\mathcal{L}_{DSA} + \lambda\,\mathcal{L}_{DIA},
\end{equation}
where $\omega$ and $\lambda$ balances the contribution of the above losses.

\section{Experiments}

\subsection{Implementation Details}
\noindent\textbf{Experiments Settings.}
Our SFDATrack is implemented using Python~3.9 and PyTorch~2.4.1. The model is trained with 2 NVIDIA A100 GPUs and tested on a single NVIDIA A6000 GPU. We employ a vanilla ViT-Base~\cite{DBLP:conf/iclr/DosovitskiyB0WZ21} network as the backbone of our tracker. The patch size is set to $16\times16$ and the feature dimension is 768. We insert three Dual Interactive Mamba (DIM) modules into the network. The template and search region sizes are set to $128\times128$ and $256\times256$, respectively. We train the model for 300 epochs and the network is optimized using AdamW with a learning rate of $4\times10^{-4}$ and a weight decay of $1\times10^{-4}$. The exponential moving average (EMA) momentum $\alpha$ is set to 0.99. The weighting coefficients for the L1, GIoU, DSA and DIA losses are set to 5.0, 2.0, 0.5 and 0.5, respectively.



\noindent\textbf{Datasets.} 
For the source-free domain adaptation training, we use three synthetic datasets: GOT-10k-Dark, GOT-10k-Foggy, and GOT-10k-Rainy generated in ~\cite{yao2025umdatrack}. Each dataset corresponds to a specific adverse condition (darkness, fog, and rain), and the video frames are sampled with a balanced sampling ratio of 1:1:1. A total of 60,000 samples are uniformly drawn for training. These datasets simulate different adverse weather scenarios and are used as unlabeled target domains. 

\begin{table}[t]
\centering
\caption{ Comparisons with state-of-the-art trackers on real-world datasets: NAT2021, UAVDark70, and AVisT. The best results are shown in bold. Trackers above the double line denote \textbf{generic} methods, and those below denote \textbf{cross-domain} methods.}
\label{tab:realworld}

\small
\setlength{\tabcolsep}{3pt}

\begin{tabular}{@{} c | cc | cc | cc @{}}
\toprule
\multirow{2.5}{*}{Tracker} & \multicolumn{2}{c|}{NAT2021} & \multicolumn{2}{c|}{UAVDark70} & \multicolumn{2}{c}{AVisT}\\
\cmidrule(lr){2-7}
& AUC & P & AUC & P & AUC & P \\
\midrule

AVTrack\cite{lilearning} & 45.41 & 59.51 & 46.91 & 59.49 & 49.21 & 48.50 \\

SMAT\cite{gopal2024separable} & 45.96 & 59.87 & 45.19 & 56.71 & 50.35 & 49.58  \\

AQATrack\cite{xie2024autoregressive} & 51.33 & 67.03 & 58.18 & 70.98 & 57.32 & 56.60 \\

SeqTrack\cite{chen2023seqtrack} & 51.65 & 67.97 & 53.88 & 66.88 & 57.15 & 55.30 \\

ROMTrack\cite{Cai_2023_ICCV} & 51.57 & 68.75 & 53.77 & 69.80 & 56.12 & 55.09 \\


EVPTrack\cite{shi2024evptrack} & 53.08 & 69.51 & 57.47 & 71.10 & 57.31 & 55.55 \\

ODTrack\cite{zheng2024odtrack} & 53.11 & 69.68 & 58.07 & 71.11 & 58.63 & 57.36 \\

ARTrackV2\cite{Bai_2024_CVPR} & 53.13 & 69.72 & 58.22 & 71.95 & 58.52 & 57.65 \\

ORTrack\cite{Wu_2025_CVPR} & 49.62 & 63.90 & 47.43 & 59.83 & 48.17 & 49.10 \\

SGLATrack\cite{DBLP:conf/cvpr/XueZLZLXS25} & 47.52 & 61.75 & 53.06 & 67.32 & 50.24 & 50.82 \\

\midrule
\midrule

MLKD-Track\cite{DBLP:journals/corr/abs-2312-07884} & 44.31 & 60.21 & 47.27 & 61.54 & 33.62 & 30.26  \\

SAM-DA\cite{Yao2023SAMDA} & 47.31 & 65.50 & 49.52 & 65.59 & 37.36 & 34.29  \\

UDAT-CAR\cite{DBLP:conf/cvpr/Ye0ZPC22} & 48.75 & 65.96 & 51.25 & 70.22 & 38.91 & 33.65  \\

DCPT\cite{DBLP:conf/icra/ZhuTCH0QLL24} & 52.55 & 69.01 & 56.86 & 70.16 & 55.66 & 52.41 \\

LVPTrack\cite{DBLP:conf/aaai/WuYHWZZR25} & 51.74 & 68.61 & 58.40 & 74.53 & 52.93 & 51.63 \\

UMDATrack\cite{yao2025umdatrack} & 54.58 & 70.78 & 60.05 & 73.35 & 60.50 & 59.01 \\

\rowcolor{gray!20} \textbf{SFDATrack} & \textbf{56.78} & \textbf{73.60} & \textbf{61.22} & \textbf{76.51} & \textbf{60.87} & \textbf{59.70} \\

\bottomrule
\end{tabular}

\end{table}

\subsection{Comparisons with State-of-the-art Trackers}
This subsection provides a comprehensive comparison of SFDAtrack with recent state-of-the-art (SOTA) trackers under diverse real-world and synthetic adverse weather conditions. Although our approach is specifically designed for cross-domain tracking, it consistently outperforms existing SOTA generic trackers, highlighting its strong generalization and robustness.


Specifically, we use synthetic datasets that include three variants from GOT-10k and DTB70, namely Dark, Foggy, and Rainy, resulting in six weather-specific domains in total. These variants are designed to simulate nighttime, foggy, and rainy environments, respectively.
In addition, two real-world datasets, NAT2021-test~\cite{DBLP:conf/cvpr/Ye0ZPC22} and UAVDark70~\cite{li2021adtrack}, are employed to evaluate tracking performance under realistic nighttime conditions. Finally, we evaluate the tracking performance on the real-world AVisT dataset \cite{DBLP:conf/bmvc/NomanGK0DDC0GK22}, which contains various challenging weather conditions in natural scenes.

\noindent\textbf{Synthetic GOT-10k and DTB70\cite{drone-tracking}.}
As shown in Table \ref{tab:comparison}, SFDATrack exhibits outstanding performance across various adverse weather conditions on both synthetic datasets. On the synthetic GOT-10k dataset, SFDATrack consistently achieves the best results across all adverse weather conditions, leading the second-best tracker by notable margins, such as 1.9\% in AO, 2.7\% in SR$_{0.5}$, and 4.9\% in SR$_{0.75}$ under dark scenarios. Similarly, on the synthetic DTB70 dataset, SFDATrack achieves the highest AUC and precision scores under almost all weather conditions.

\begin{table}[t]
\centering
\caption{Comparison of inference speed, MACs, and model parameters across different trackers.}
\label{tab:comparison_fps}
\small

\begin{tabular}{c|c|c|c} 
\toprule
Tracker & Speed (FPS) & MACs (G) & Parameters (M) \\ 
\midrule

SeqTrack\cite{chen2023seqtrack} & 40 & 66 & 89 \\

DropTrack\cite{dropmae2023} & 52 & 48 & 92 \\

AQATrack\cite{xie2024autoregressive} & 68 & 26 & 72 \\

EVPTrack\cite{shi2024evptrack} & 71 & 22 & 74 \\

ARTrackV2\cite{Bai_2024_CVPR} & 95  & 45 & 126  \\

DCPT\cite{DBLP:conf/icra/ZhuTCH0QLL24} & 30 & 30 & 95 \\


\rowcolor{gray!20} \textbf{SFDATrack} & \textbf{91} & \textbf{30} & \textbf{93} \\

\bottomrule
\end{tabular}
\end{table}

\begin{table}[b]
\begin{minipage}[t]{0.49\columnwidth}
\centering
\caption{ Ablation study on the individual impact of each module (DIM, HPP) in our model. The presence or absence of each module is marked with a check or dash, respectively. The performance is evaluated on the NAT2021 dataset.}
\label{tab:effectiveness_metrics}

\small

\begin{tabular}{@{}cc|cc@{}} 
\toprule
\multicolumn{2}{c|}{\textbf{Modules}} & \multicolumn{2}{c}{\textbf{Indicators}} \\  
\midrule
DIM & HPP & AUC (\%) & Precision (\%) \\
\midrule
- & - & 51.22 & 67.62 \\
$\checkmark$ & -  & 53.56 & 70.88 \\
- & $\checkmark$ & 53.24 & 70.37 \\
$\checkmark$ & $\checkmark$ & \textbf{56.78} & \textbf{73.60}  \\
\bottomrule
\end{tabular}
\end{minipage}
\hfill
\begin{minipage}[t]{0.49\columnwidth}
\centering
\caption{Ablation study on different alignment strategies within the HPP module. The compared variants include Domain-Specific Alignment (DSA) and Domain-Invariant Alignment (DIA). Results are evaluated on the UAVDark70 dataset.}
\label{tab:HPP_ablation}

\small

\begin{tabular}{@{}cc|cc@{}} 
\toprule
\multicolumn{2}{c|}{\textbf{Modules}} & \multicolumn{2}{c}{\textbf{Indicators}} \\  
\midrule
DSA & DIA & AUC (\%) & Precision (\%) \\
\midrule
- & - & 58.29 & 72.37 \\
$\checkmark$ & -  & 59.65 & 74.07 \\
- & $\checkmark$ & 59.55 & 73.91 \\
$\checkmark$ & $\checkmark$ & \textbf{61.22} & \textbf{76.51}  \\
\bottomrule
\end{tabular}
\end{minipage}
\end{table}

\noindent\textbf{Results on Real-World Datasets}
Besides achieving outstanding performance on synthetic datasets, we further evaluate SFDATrack on real-world benchmarks to verify its generalization ability in practical scenarios. As shown in Table~\ref{tab:realworld}, SFDATrack exhibits superior performance across all three real-world datasets. On the NAT2021 dataset, it achieves the highest AUC and precision scores of 56.78 and 73.60, respectively, outperforming the second tracker UMDATrack by 2.20\% and 2.82\%. On UAVDark70 dataset, SFDATrack again achieves the best AUC (61.22) and precision (76.51), surpassing the previous best tracker by 1.17\% and 3.16\%. Furthermore, on the challenging AVisT dataset, which involves various adverse weather and illumination conditions, 
SFDATrack attains the best AUC (60.87) and precision (59.70), clearly outperforming other advanced trackers. 
These results demonstrate the strong generalization ability and robustness of SFDATrack when applied to real-world tracking scenarios.

\begin{table}[t]
\centering
\begin{minipage}[t]{0.49\columnwidth}
\centering
\caption{Ablation study on the impact of different architectures within the DIM module. The compared variants include Transformer, Uni-Mamba, Bi-Mamba, and our full DIM module. The results are evaluated on the UAVDark70 dataset.}
\label{tab:dim_ablation}

\small

\begin{tabular}{@{}c|@{\hspace{3pt}}cc@{}}
\toprule
Architecture & AUC (\%) & Precision (\%) \\
\midrule
Transformer~\cite{vaswani2017attention} & 56.77 & 70.45 \\
Uni-Mamba~\cite{mamba} & 58.85 & 74.58 \\
Bi-Mamba~\cite{DBLP:conf/icml/ZhuL0W0W24} & 60.33 & 75.60 \\
\textbf{DIM (ours)} & \textbf{61.22} & \textbf{76.51} \\
\bottomrule
\end{tabular}
\end{minipage}
\hfill
\begin{minipage}[t]{0.49\columnwidth}
\centering
\caption{Experiments on the weighting coefficients of Domain-Specific Alignment (DSA) $\omega$ and Domain-Invariant Alignment (DIA) $\lambda$. Results are reported in terms of AUC and Precision on the NAT2021 dataset.}
\label{tab:analyze_weight}


\begin{tabular}{c@{\hspace{10pt}}c@{\hspace{5pt}}|@{\hspace{3pt}}cc}
\toprule
$\omega$ & $\lambda$ & AUC (\%) & Precision (\%) \\
\midrule

0.1 & 0.1 & 56.16 & 73.20\\

0.1 & 0.5 & 55.09 & 71.67  \\

0.5 & 0.1 & 55.37 & 72.12  \\

\textbf{0.5} & \textbf{0.5} & \textbf{56.78} & \textbf{73.60} \\
\bottomrule
\end{tabular}
\end{minipage}

\end{table}

\begin{table}[h]
\centering
\caption{The study on the update frequency of EMA (Exponential Moving Average).}
\label{tab:ema_frequency}
\small

\begin{tabular}{@{}c|cc@{}}
\toprule
EMA Frequency  & AUC (\%)  & Precision (\%) \\
\midrule
Each batch      & 55.20 & 72.47 \\
Every 5 epochs  & 54.32 & 70.41 \\
Each epoch      & \textbf{56.78} & \textbf{73.60} \\
\bottomrule
\end{tabular}
\end{table}

\noindent\textbf{Inference Speed.} As demonstrated in Table \ref{tab:comparison_fps}, SFDATrack achieves a good balance between accuracy and efficiency.  Compared with existing state-of-the-art trackers, SFDATrack achieves a similar inference speed to ARTrackV2, while requiring fewer parameters and lower computational cost. Moreover, SFDATrack runs notably faster than most other trackers, such as SeqTrack (40 FPS) and DropTrack (52 FPS), demonstrating its superior efficiency and real-time capability.

\subsection{Ablation Studies}

\noindent\textbf{Study on the Components of SFDATrack.} We conducted ablation experiments on the proposed two modules to verify their effectiveness. As shown in Table \ref{tab:effectiveness_metrics}, when the DIM module is introduced, the model achieves an AUC of 2.34 points and a Precision of 3.26 points higher than the baseline. When adding the HPP module, it improves the AUC to 53.24 and the Precision to 70.37. While both the DIM module and HPP module are introduced, it achieves the best performance with an AUC of 56.78 and Precision of 73.60. These results indicate that each module plays an important role in the whole model.

\noindent\textbf{Study on the Architecture of HPP.}
We further conduct ablation experiments to validate the effectiveness of the proposed HPP module. As shown in Table \ref{tab:HPP_ablation}, when the Domain-Specific Alignment (DSA) is introduced, the AUC and Precision are improved by 1.36\% and 1.70\%, respectively. And when the Domain-Invariant Alignment (DIA) is introduced, the AUC and Precision are improved by 1.26\% and 1.54\%. However, when both DSA and DIA strategies are incorporated, the model achieves the best performance with an AUC of 61.22 and a Precision of 76.51 on the UAVDark70 dataset.

\noindent\textbf{Study on the Architecture of DIM.}
We investigate the effect of different architectural designs within the DIM module. As shown in Table~\ref{tab:dim_ablation}, replacing the Transformer~\cite{vaswani2017attention} with the Mamba-based~\cite{mamba} structure brings a clear performance gain. Specifically, Uni-Mamba (Unidirectional Mamba) achieves an improvement of 2.08\% in AUC and 4.15\% in Precision over the Transformer baseline. Introducing bidirectional modeling in Bi-Mamba~\cite{DBLP:conf/icml/ZhuL0W0W24} further improves AUC by 1.48\%  and Precision by 1.18\% over the Uni-Mamba. Finally, our full DIM achieves the best results in the UAVDark70 dataset, with AUC of 61.22 and Precision of 76.51, confirming the effectiveness of the proposed bidirectional Mamba design.

\begin{figure*}[t]
  \centering
  \includegraphics[width=\textwidth]{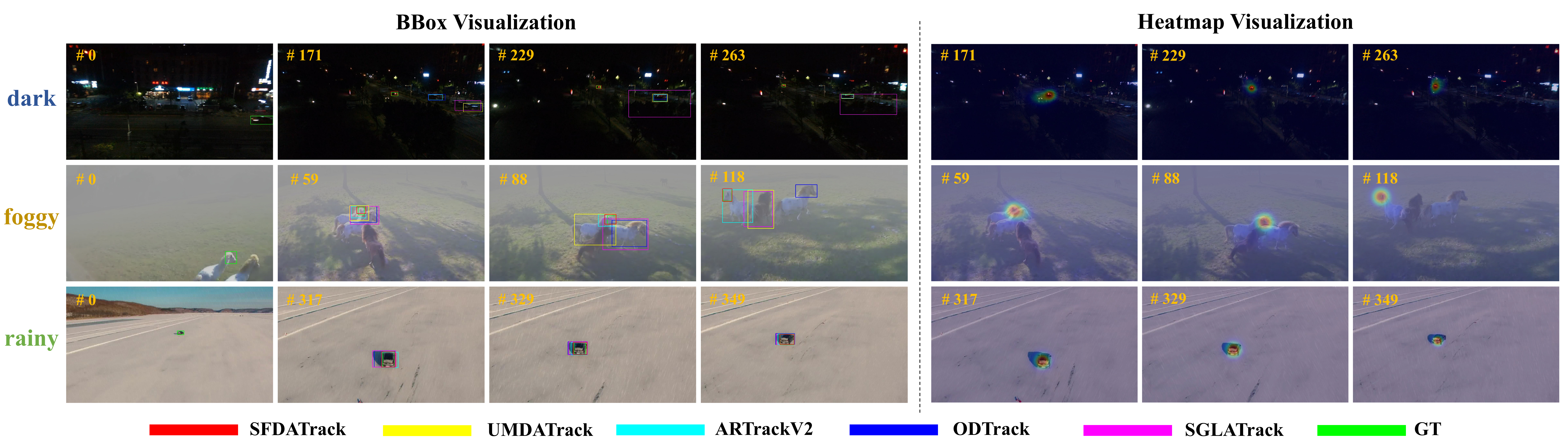}
  \caption{Visualization results of SFDATrack under three challenging weather conditions: dark, foggy, and rainy scenes. The left part shows the bounding box (BBox) visualizations, while the right part presents the corresponding heatmaps.}
  \label{fig:bbox}
\end{figure*}

\noindent\textbf{Study on Training Hyperparameters.}
We further analyze the loss weights of Domain-Specific Alignment (DSA) and Domain-Invariant Alignment (DIA), as well as the EMA update frequency. As shown in Table~\ref{tab:analyze_weight}, a balanced weighting between DSA and DIA (0.5, 0.5) performs best, while imbalanced settings degrade performance, indicating that both terms are necessary for effective adaptation. Moreover, Table~\ref{tab:ema_frequency} indicates that updating EMA once per epoch yields the most favorable performance.



\subsection{Visualization Results and Generalized Ability}
\noindent\textbf{Visualization in Adverse Conditions.}
Fig. \ref{fig:bbox} illustrates the qualitative tracking results, where SFDATrack exhibits higher accuracy and stronger robustness than other trackers under extreme weather conditions. Even in real nighttime scenarios with small, barely visible targets, SFDATrack is still able to localize the object clearly.
The corresponding heatmaps further demonstrate that SFDATrack focuses more precisely on target regions across frames, maintaining concentrated attention in challenging scenes. 

\begin{figure}[htb]
  \centering
  \includegraphics[width=4in]{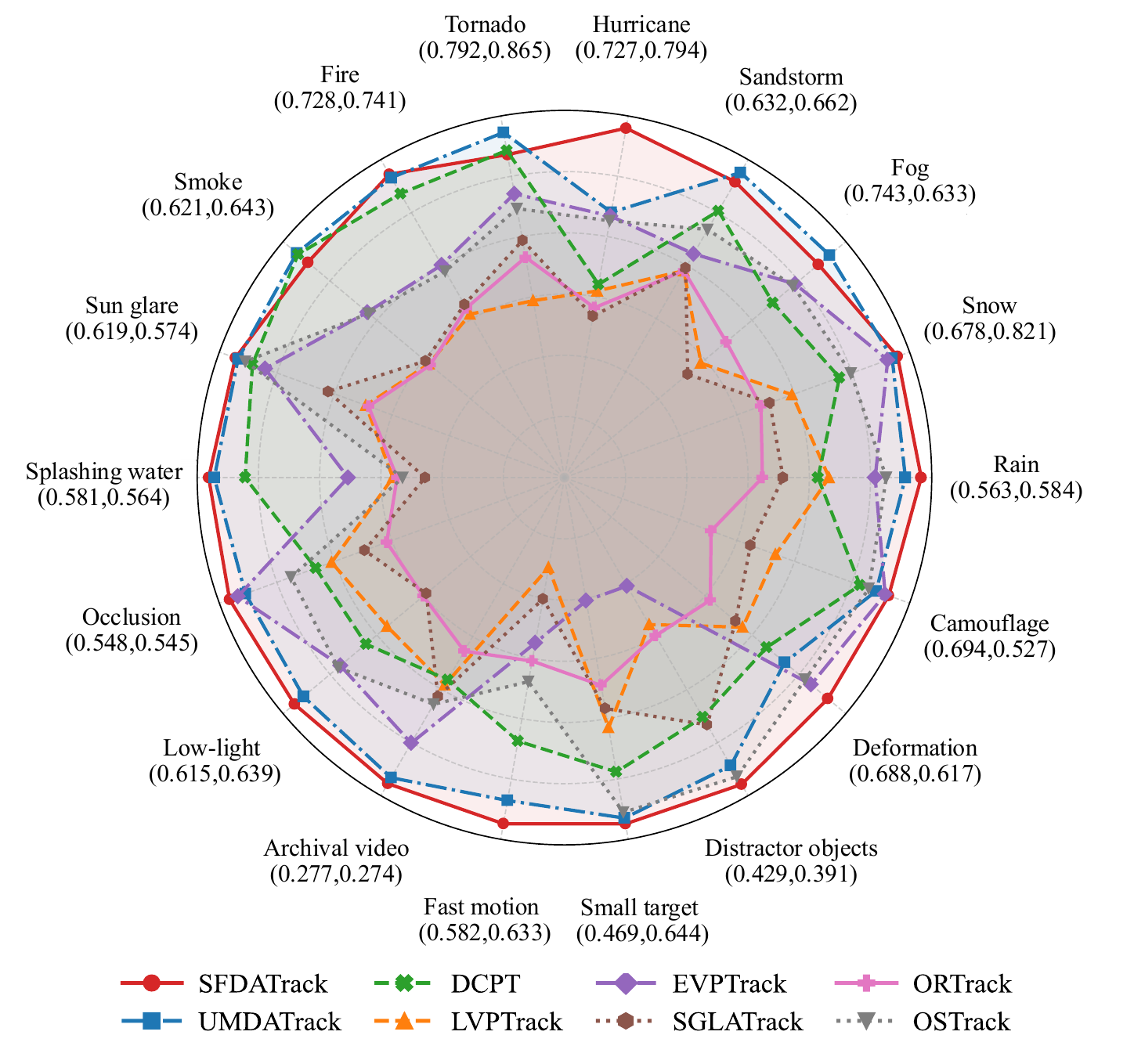}   
   \caption{The performance of our method compared with other trackers in terms of AUC and Precision on eighteen attributes of the real-world AVisT dataset. Each axis corresponds to one attribute, with the radius representing the AUC value under the corresponding condition.}
  \label{fig:radar}
\end{figure}

\noindent\textbf{Generalized Ability.}
As shown in Fig. \ref{fig:radar}, we further analyze the performance of SFDATracker on different challenging attributes of the AVisT dataset. Specifically, SFDATrack achieves the best performance across most adverse weather scenarios and challenging attributes, such as hurricane, rain, snow, low-light, fast motion, among others. For several other challenging conditions, such as fog and sandstorm, it achieves the second-best performance, demonstrating strong robustness and generalization ability under diverse adverse visibility settings.

\section{Conclusion}
\label{sec:conclusion}

In this work, we introduce a generalized source-free domain adaptive tracker termed SFDATrack that only leverages adverse weather samples from the target domain for robust state estimation. We propose a Dual Interactive Mamba module that uses bidirectional sequence interaction and prompt token exchange strategies to narrow domain discrepancies. Furthermore, we design a Hyperspherical Prototype Projection module to project tokens onto multi-domain prototypes within a latent hyperspherical space, enforcing both domain-specific and domain-invariant properties for model generalization. Extensive experiments on multiple adverse weather datasets demonstrate the effectiveness of the proposed SFDATrack.

\section{Acknowledgments}
\label{sec:acknowledgments}

This work was supported in part by the National Natural Science Foundation of China (Grant No. 62402055, 62302053, 62322216, and U2541229) and the Shenzhen Science and Technology Program (Grant No. KQTD20221101093559018 and SYSRD20250529113401002).
%
%
\bibliographystyle{splncs04}
\bibliography{main}
\end{document}